# The complexities of patient-centred conversational artificial intelligence

João Matos[1,2*], Olivia Buege[1], Donny Cheung[1], Gary S. Collins[3,4], Paula Dhiman[2], Nan Li[1], Bingyu Mao[1], Benjamin W. Nelson[1,5], Michail Ouroutzoglou[1], Paul Varghese[1], Jonathan Amar[1]

[1] Verily Health, Dallas, TX, USA

[2] Centre for Statistics in Medicine, University of Oxford, Oxford, UK

[3] Department of Applied Health Sciences, School of Health Sciences, University of Birmingham, Birmingham, UK

[4] National Institute for Health and Care Research (NIHR) Biomedical Research Centre: Birmingham, Birmingham, University Hospitals Birmingham NHS Foundation Trust and University of Birmingham, Birmingham, UK

[5] Division of Digital Psychiatry, Department of Psychiatry, Harvard Medical School and Beth Israel Deaconess Medical Center, Boston, MA, US

* Corresponding Author, joao.matos@ndorms.ox.ac.uk

## Abstract

Consumer-facing health chatbots powered by large language models (LLMs) are increasingly used for symptom assessment. However, chatbot development and evaluation often rely on cooperative, articulate, simulated patients. We analysed 2,053 real patient-chatbot conversations and found that communication patterns and expression of emotions vary widely across users. We developed a patient simulator that separately models clinical content, emotional state, conversational strategy, and communication style. In a Turing-inspired evaluation of realism with 15 human graders, simulated conversations were nearly indistinguishable from real ones, with human graders achieving an accuracy of 55%. We used five distinct patient personae, across 1,164 clinician-graded cases, to evaluate the performance of four LLMs in urgency assessment. We found that communication style can significantly alter triage outcomes. Patient-centred conversational artificial intelligence must accommodate communication diversity: systems designed for idealised, rather than realistic, interactions risk underperforming and amplifying health disparities when deployed in the real world.

Conversational artificial intelligence (AI) is becoming an important interface between patients and health care [1]. Consumer-facing applications with health assistants powered by large language models (LLMs), including ChatGPT Health [2], *Verily Me* [3], Microsoft Copilot Health [4], and, more recently, Google Health [5] allow individuals to describe symptoms, receive preliminary assessments, and decide when to seek care.

What these human-chatbot conversations gain in convenience and cost reduction is arguably lost in the wealth of contextual patient information. A physician conducting an in-person consultation draws on visual cues, body language, vocal tone, and physical examination findings that text-based exchange cannot fully capture [6]. Multimodal inputs such as images [7,8] , sensor data from wearable devices [9], or synchronised electronic health records [10] can only partially compensate, hardly substituting for the contextual richness of direct clinical encounter.

Consumer-facing medical AI differs fundamentally from clinician-facing AI. In clinician-facing systems, such as imaging decision-support tools, a trained professional remains responsible for interpreting the output and integrating it into their clinical workflow [11]. In consumer-facing applications, such as health chatbots, the system interacts directly with patients. The performance of the system will also depend on *what* patients understand and disclose, and *how* they communicate. The onus now falls more heavily on the patient: often without clinician supervision, users themselves must judge whether they are receiving safe, useful, and fair guidance.

This makes conversational medical AI uniquely complex. Emotional, linguistic, and literacy nuances mediate every exchange. For instance, stigma or embarrassment may make a patient omit a relevant symptom unless the system elicits it sensitively. An anxious user may inadvertently bury a critical detail within a flood of unrelated questions, diverting the conversation away from the relevant clinical issue. A patient with limited English proficiency may describe nasal congestion as "cannot take air", potentially leading a model that lacks linguistic and cultural nuance to infer a respiratory emergency. Users with low digital literacy may abandon the session before receiving a recommendation, and therefore never appear in datasets of completed consultations. A human clinician may recognise and adapt to such patterns; a chatbot will do so only if it is explicitly designed and evaluated for them.

Conversational AI is therefore exposed to missing information, distributional shifts, and sample selection bias, as with traditional machine learning systems [12]. In this new setting, these problems arise from the patient-system interaction itself, with the digital divide shaping who is affected and how [13].

These complexities are rarely captured in current AI benchmarks [14]. LLMs pass medical licensing examinations with high scores[15], generate clinical summaries rated equivalent to or better than those written by physicians [16], and achieve strong diagnostic accuracy on challenging clinical cases [17]. These results have driven optimism about clinical deployment, but they do not necessarily translate to increased patient benefit [18]. Bean et al. found that participants seeking health advice using LLMs identified relevant conditions and chose appropriate dispositions at rates no better than a control group using simple internet search [19]. When the human participants were replaced by LLM-simulated patients, system accuracy returned to benchmark levels. The models performed better with synthetic users than with real ones, a symptom of a deeper problem: the design of **medical AI is rarely centred on the patient or the interactional contexts of intended use** [20,21].

Real patient interaction for system development is slow, costly, and raises privacy questions [1]. It is also difficult (if not impossible) to construct a fully reproducible conversational dataset: LLM outputs are stochastic, and multi-turn clinical conversations are dynamic, path-dependent interactions. Even when real interactions are collected, annotation at scale faces considerable economic barriers [22]. Human studies often produce samples that are too small for adequate model evaluation [23], or are demographically narrow to support generalisable conclusions [24].

These difficulties explain the use of LLM-based patient simulations to test conversational medical AI systems at scale [25]. In this approach, an AI system role-plays a patient during a multi-turn clinical encounter, allowing automated evaluation of AI clinicians, triage systems, and health coaches.

Despite their growing use, patient simulators are rarely subjected to the same scrutiny as the systems they are intended to evaluate [26]. There is limited evidence that LLMs are accurate simulators of real patient dialogue, internally consistent with their own clinical presentation [27], or representative of the diversity of human behaviour. Wang et al. argue that LLMs are potentially harmful substitutes for human participants in identity-relevant research because they tend to misrepresent and homogenise demographic groups in ways that reproduce epistemic injustice against marginalised communities [28,29]. In patient simulation, this concern is especially pertinent: current simulators often generate cooperative, articulate, and literate responses regardless of the clinical scenario or patient profile [1].

How a patient communicates, through verbosity, choice of vocabulary, goal-directed behaviour, or emotional expression, is a dimension of the clinical interaction that cooperative simulators flatten and evaluation benchmarks rarely vary. Whether the same clinical presentation, communicated differently, yields different guidance has not been studied.

Our objective was to evaluate the effect of patient-side conversational patterns on the performance of conversational AI. Evaluating these effects required a patient simulator capable of reproducing them. In this paper, we detail our progress toward the design of a modular, tunable LLM-based patient simulator. We were inspired by the "three conversations model" [30], which argues that three parallel conversations take place during difficult exchanges, such as clinical consultations. With this architecture, we address patient simulation by explicitly and separately modelling patient's *feelings* and *strategy*, in addition to clinical *content* (facts). The design was grounded in an initial analysis of real patient conversations drawn from the *Verily Me* mobile application. We evaluated the patient simulator along several dimensions, including parameter adherence, clinical fidelity, and realism. Realism was assessed in a Turing-inspired trial [31] with both human and LLM graders. We then applied the simulator to assess the performance and fairness of LLM-based urgency assessment, one of the top three use cases of public-facing health LLMs [2,4]. Clinical cases were constructed for urgency assessment and graded by two to five clinicians each. For each case, distinct personae were developed to reflect varied communication profiles, allowing evaluation of how the different conversational styles affected the performance of different LLM-based medical systems.

## Results

### Real patient-AI conversations are complex and heterogeneous

Among 2,053 patient-AI conversations that took place in the *Verily Me* mobile application, 289 (14%) reached a completed triage recommendation (median 17 patient turns), 177 (9%) were terminated early by an emergency flag (median 7 turns), and 1,587 (77%) were abandoned by the user before completion (median one turn) (**Figure 2a**). Further, across the 9,196 patient messages in the 1,006 evaluable sessions (those with at least two patient turns), message length varied widely, with 67% of patient responses consisting of fewer than six words (**Figure 2b**).

Symptom categories spanned 14 categories, including 9% genitourinary, 8% neurological, 8% musculoskeletal, and 7% gastrointestinal complaints (**Figure 2c**). Health literacy was predominantly functional/basic (60%) and communicative/medium (39%), with < 1% of conversations rated as critical/high (**Figure 2d**). Patient recall of facts was high in 8% of conversations, medium in 63%, and low in 29%. Patients volunteered information beyond what was asked in 30% of the sessions.

Emotional signals were also common: 37% of conversations contained at least one (**Figure 2e**). Anxiousness was the most frequent emotional signal (21% of the sessions exhibited it), followed by frustration (19%), scepticism (7%), impatience (4%), and embarrassment (3%). Non-standard communication features appeared in 79% of the sessions (**Figure 2f**), including non-standard capitalisation (23%), missing punctuation (52%), grammar issues (64%), and typographic errors (43%). These patterns varied across individuals and often occurred simultaneously.

### Realistic patient simulation

We developed a patient simulator with a multi-node architecture that decomposes patient behaviour into four processing channels: clinical content, emotional state, conversational strategy (in parallel), and communication style (**Figure 1a**). Each channel is a separate LLM call. In total, 20 parameters can be independently adjusted to shape the simulated patient's behaviour (**Figure 1b**). The full architecture, parameters, and implementation are described in the **Methods**. We evaluated the simulator along three dimensions: parameter adherence, clinical fidelity, and realism (**Figure 1c-e**).

#### *Parameter adherence*

Of the 20 simulation parameters, 17 were assessed with a comparative discrimination protocol that tested whether each produces a measurable effect on the simulator's output, and the remaining three (verbosity and two session-termination behaviours) with deterministic compliance checks. Across 950 pairs of conversations, the activated conversation was correctly identified in 847 cases (macro concordance statistic = 0.894; **Figure 3a**). Seven parameters achieved perfect discrimination (c-statistic = 1.0): sensitive topics, capitalisation, communication style, English proficiency, punctuation style, anxiousness, and scepticism. Patient goal (0.98), embarrassment and frustration (both 0.94), and health literacy (0.92) were close behind. Oversharing (0.68) and typos (0.60) were the weakest, likely because these may have a sparser presence on the simulation side to avoid overly formulaic text. For the three deterministically checked parameters, session-termination behaviours were followed in 100% of cases, and increasing verbosity levels (*minimal* > *very brief* > *brief* > *moderate* > *verbose*) translated to an increasing average number of words per conversation (**Supplementary Material 3**).

*Clinical Fidelity*

Across 50 conversations, 442 supported claims achieved a mean accuracy of 3.94/4, with 94.3% rated fully accurate and no (0) contradictions (**Figure 3b**). The 238 unsupported claims achieved a mean plausibility score of 3.79/4. The patient simulator generated accurate statements and clinically reasonable elaborations when asked about details absent from the source material (**Supplementary Material 4**).

*Realism*

Human graders achieved an overall accuracy of 55.0% (95% confidence interval (CI): 47.0, 62.8), a simulation detection rate of 55.4% (95% CI: 44.1, 66.2), and a real recognition rate of 54.7% (95% CI: 43.4, 65.4). Among LLM judges, Claude Opus 4.6 achieved an overall accuracy of 81.9% (95% CI: 75.2, 87.1), simulated detection rate of 70.0% (95% CI: 59.2, 78.9), and a real recognition rate of 93.8% (86.2, 97.3). GPT-5.4 showed an overall accuracy of 61.9% (95% CI: 54.2, 69.0), a simulated detection rate of 25.0% (16.8, 35.5), and a real recognition rate of 98.8% (95% CI: 93.3, 99.8), a consequence of classifying 87% of conversations as real **(Figure 3c**). **Supplementary Material 5** provides protocol details, sample size estimations, and further analysis.

**Communication style impacts urgency assessment by AI**

We used the patient simulator to evaluate whether patient communication style impacts the performance of LLM-based urgency assessment (**Figure 1f**). In total, 1,164 clinician-graded clinical cases (labelled for urgency assessment by between two to five clinicians) were simulated under five personae: *Anxious Patient*; *Dismissive Patient; Informed Advocate*; *Limited Communicator*; and a *Default* baseline. These were evaluated across four LLM-based clinician models, Gemini 3.5 Flash (showcased in **Figures 4 and 5**), GPT-5.5, GPT-5.4-mini, and Claude Opus 4.6 (**Extended Data Table 1**; fully documented in **Supplementary Material 7,** including computational costs).

*Over-triage and under-triage*

For Gemini 3.5 Flash, the proportion of over-triage was 36.8% (95% CI: 33.1, 40.4) for the *Anxious Patient* persona, 32.0% (95% CI: 28.7, 35.5) for the *Informed Advocate*, 30.8% (95% CI: 27.5, 34.1) for the *Limited Communicator*, 25.8% (95% CI: 22.6, 29.0) for the *Default* baseline, and 23.3% (95% CI: 20.4, 26.4) for the *Dismissive Patient*. The proportion of under-triage followed the inverse ordering: 2.5% (95% CI: 1.6, 3.5) for the *Anxious Patient*, 4.3% (95% CI: 3.0, 5.6) for the *Limited Communicator*, 4.5% (95% CI: 3.2, 5.8) for the *Informed Advocate*, 5.6% (95% CI: 4.1, 7.1) for the *Default* baseline, and 6.6% (95% CI: 5.1, 8.2) for the *Dismissive Patien*t. Over-triage was computed among the 721 cases per persona whose label was not urgent, and under-triage among the 964 whose label was not self-care. Performance metrics are available in **Extended Data Table 1**.

In pairwise performance comparisons, patient persona significantly altered urgency assessment by LLMs (**Figure 4**). For Gemini 3.5 Flash, against the *Default* baseline, over-triage rose by 11.0 percentage points (pp) under the *Anxious Patient* persona (95% CI: 8.3, 13.7); 6.2 pp (95% CI: 3.6, 8.9) under the *Informed Advocate*; 5.0 pp (95% CI: 2.5, 7.6) under the *Limited Communicator*, and fell by 2.5 pp (95% CI: 0.1, 4.9) under the *Dismissive Patient* persona. When comparing the *Anxious Patient* persona with the *Dismissive Patient* persona, the over-triage rate difference reached 13.5 pp (95% CI: 10.5, 16.4). Under-triage, the more clinically concerning error, moved in the opposite direction. Relative to baseline, it fell by 3.1 pp (95% CI: 2.0, 4.4) under the *Anxious Patient persona* and edged down under the *Informed Advocate* by 1.1 pp (95% CI: 0.0, 2.3) and *Limited Communicator* by 1.3 pp (95% CI: 0.0, 2.7), while rising by a non-significant 1.0 pp (95% CI: -0.3, 2.4) under the *Dismissive Patient* persona. The *Anxious*-versus-*Dismissive* contrast was 4.1 pp (95% CI: 2.8, 5.6). **Box 1** illustrates this pattern.

The direction of these effects was consistent across the other models, albeit with smaller magnitudes. The over-triage difference between *Anxious* and *Dismissive* personae was 8.7pp for GPT-5.5, 8.2pp for GPT-5.4-mini, and 5.4pp for Claude Opus 4.6. Under-triage differences followed the same ordering: 3.6pp for GPT-5.5, 4.1pp for GPT-5.4-mini, and 1.3pp for Claude Opus 4.6 (**Supplementary Material 7**).

*Discrimination and calibration*

To examine the mechanism behind persona-driven urgency assessment differences, we estimated discrimination and calibration for models using multiple LLM assessment repeats. **Figure 5** shows results on Gemini 3.5 Flash; GPT models can be found in **Supplementary Material 7;** Claude Opus 4.6 was excluded from this analysis because it does not have a seed hyperparameter. Concordance statistics (c-statistic) were similar across personae for all three one-versus-one class comparisons (**Figure 5a, c, e**). The models ranked cases with comparable discrimination regardless of how the patient communicated. Calibration-in-the-large, by contrast, differed substantially. All personae showed over-estimation of P(Urgent), i.e., the extracted urgency probabilities were systematically higher than the observed prevalence, but the magnitude of this over-estimation varied by persona. For Gemini 3.5 Flash, the *Anxious* persona produced the largest over-estimation of P(Urgent), at +12.4 (95% CI: 10.4, 14.4) pp, and the largest under-estimation of P(Self-Care), at -9.6 (95% CI: -11.5, -7.8) pp. On the other hand, the *Dismissive* persona showed the smallest miscalibration in both directions: -4.3 (95% CI: -6.2, -2.4) pp for P(self-care) and +5.1 (95% CI: 3.5, 6.8) pp for P(urgent).

*Demographic subgroup analysis*

The proportion of triage errors stratified by patient sex and age group showed overlapping confidence intervals across all subgroups for each persona (**Extended Data Figure 1**). Over-triage and under-triage patterns were broadly consistent between female and male patients and across age groups (≤30, 31-50, 51-70, 71+). The persona effect, with higher over-triage for anxious patients, higher under-triage for dismissive patients, was present in the same direction within every demographic stratum. The analysis was not powered to detect interactions between persona and demographic group; subgroup sample sizes ranged from 91 (age 71+ for over-triage) to 593 (female for under-triage), and the study was designed to detect persona-level differences, which were estimated to require 1,200 samples (full-cohort) (see **Supplementary Material 7** for further details on sample size estimation).

## Discussion

We set out to understand whether the *way* a patient communicates, separate from *what* they communicate, changes the behaviour of conversational AI. Addressing this question required a simulator capable of holding clinical facts fixed while varying communication styles. We analysed 2,053 real patient-AI conversations, built a patient simulator, and thoroughly evaluated it. We then used it to evaluate four LLM-based clinician models across 1,164 clinician-graded cases, presenting each case under five communication personae.

In the real conversations we analysed, patients wrote short messages, abandoned sessions early, expressed emotion, and rarely used the polished prose that LLMs default to. Only 23% of sessions reached an urgency assessment and recommendation. A symptom checker that requires sustained, articulate engagement to deliver value may fail users who most need a low-friction route to care. Engagement, brevity, and the ability to extract a useful recommendation from a sparse and imperfect exchange are core design and safety requirements for conversational medical AI.

The patient simulator performs well on parameter adherence and clinical fidelity, which is unsurprising given the capabilities in instruction-following and factual consistency of current LLMs. Even though the simulator is able to capture a wide behavioural range, it remains a coarse approximation of exactly how people talk to chatbots. Building a more accurate model will need further empirical account of patient-chatbot behaviour, which in turn needs interdisciplinary work with clinicians, social scientists, and the patients themselves. We did not test how well the simulator represents specific demographic or cultural groups [28]. Our data, graders, and co-authors are anchored in the English-speaking world (US, Canada, and UK).

The realism evaluation suggests that simulated conversations can pass for real, within limits. Human graders distinguished simulated from real conversations at 55% accuracy, which is close to random. The limitations of simulation are visible in Claude Opus 4.6's performance, which detected simulated patients more reliably than humans did, picking up on prose regularities that people miss. A single simulator, however well-tuned, produces a narrower band of behaviour than a population does. Grader free-text comments pointed at certain artefacts repeatedly, including typos that were too consistent and answers that stayed too neatly on topic. The number of words appears to have been used as a shortcut for most human graders, with longer messages being more likely to be classified as simulated.

The central finding of this study is that patient communication style impacts LLMs' urgency assessment for identical clinical content. For example, "anxious presentations" raised over-triage and lowered under-triage; "dismissive presentations" showed the reverse, and the pattern held across all four LLM-based clinician models. The other two personae, which varied health literacy and language proficiency rather than emotional tone (the *Informed Advocate* and *Limited Communicator*), shifted urgency assessment to a lesser extent, but both still tended to raise over-triage relative to the Default baseline. This is not unexpected: human clinicians are likewise influenced by how a patient presents, and in LLMs the effect is consistent with documented sycophancy, i.e., the tendency to align responses with the view a user expresses[32]. This sensitivity to communication style nonetheless has a direct consequence for how the performance of conversational AI is reported. Benchmarks built on cooperative, literate, and articulate simulated patients measure a best case; consequently, reported performance is likely inflated relative to what real users would obtain.

Assessing discrimination and calibration attempted to understand the mechanism driving the models' differences in urgency assessment. Discrimination was similar across personae (as measured by c-statistics), whereas calibration (as measured by calibration-in-the-large) shifted. This means that the models ranked cases similarly but moved their "decision thresholds" according to how the patient sounded. If the disparity originates mainly in calibration, recalibration conditioned on communication features could reduce it, though this presumes those features can be measured reliably and safely acted upon at inference time. Treating an LLM urgency assessment system as a clinical prediction model and examining it in terms of discrimination and calibration was informative here. As estimated probabilities could not be read from the model directly, since no established method exists for doing so, we reconstructed each one from repeated runs of the final conversation turn.

A growing body of literature has documented that LLMs produce different clinical recommendations when patient vignettes are labelled with different sociodemographic attributes [33]. Cases labelled as Black receive more aggressive triage, and LGBTQIA+ subgroups are recommended mental health assessments at rates several times the physician-derived baseline [34]. These counterfactual analyses hold clinical details constant and vary only demographic identifiers, testing for explicit bias in model outputs [35,36]. As AI guardrails improve, overt demographic differences may diminish [2]. Communication style is a less visible axis that we argue to deserve equal scrutiny. As our findings demonstrate, altering a patient's communication style moved urgency assessment performance by as much as 13.5 percentage points for identical clinical cases.

The effect was not uniform across the dimensions along which patients differed. Personae defined by emotional presentation (Anxious and Dismissive) produced the largest and most consistent shifts, whereas the personae defined by health literacy and language proficiency (Informed Advocate and Limited Communicator) did not differ significantly from one another. Verbosity, vocabulary, and emotional expression are profoundly shaped by an individual's linguistic background [37], health and digital literacy [38–40], and socioeconomic status, all of which are social determinants of health and well-known drivers of existing healthcare inequalities [41]. Current consumer-facing AI shifts the burden onto the patient to communicate in a way that yields a safe and useful recommendation, and not everyone is equally equipped to do so. This should be addressed mainly through inclusion of better and more equitable system designs, not by simply expecting patients to improve their own health and digital literacy. A system built to serve the communication styles currently least served by these tools, instead of an idealised default, would likely give safer and clearer guidance to those users and everyone else.

Our study has several limitations. We examined one clinical task (urgency assessment), with a single three-level outcome; other tasks and open-ended outcomes (such as recommendations) may behave differently. Although every case was annotated by two to five clinicians, a simulated conversation that introduces new clinical facts, compared to its source vignette, could weaken the link between a case and its reference label. We report clinical fidelity performance of the patient simulator to bound this risk. Several of our evaluations also rely on LLM-as-a-judge (analysis of real conversations, parameter adherence, clinical fidelity, and part of realism evaluation), which we manually inspected as needed. The real conversations that motivated and grounded our simulator also carry an inherent selection bias: as users of a digital health tool, these patients likely possess a baseline of digital literacy and smartphone access, which may not reflect the communication styles of populations on the other side of the digital divide. Additionally, running 1,164 cases across five personae and four models across multi-turn conversations is computationally expensive, which constrained how many personae and models could be explored; the four persona-archetypes here sample a small part of a large space. We evaluated only models served through a commercial application programming interface (API), not open-weight models. Behaviour is likely to change as models are updated, so any single snapshot evaluation such as this one is time-limited [42] and these evaluations need to be repeated over time.

Methods for representing patients in the evaluation of medical AI have received far less attention and scrutiny than the conversational AI systems being evaluated. We argue that patient-centred design belongs at the core of consumer-facing conversational AI. Furthermore, as recursive self-improvement of AI agents gains traction [43], it has been argued that more sophisticated simulation environments are needed for dynamic evaluation of AI [25]. Prospective evaluation with real users at scale would settle questions that simulation can only approximate, but it carries financial, time, and ethical costs [22]. Patients interacting with a system under evaluation may not benefit equally, and could be harmed in time-critical situations. Simulation, used carefully, offers a way to study these effects before deployment. The framework presented here is an early step that we hope will shape how the community develops and evaluates conversational medical AI.

## Exhibits

### **Figure 1.** Overall framework and methodology

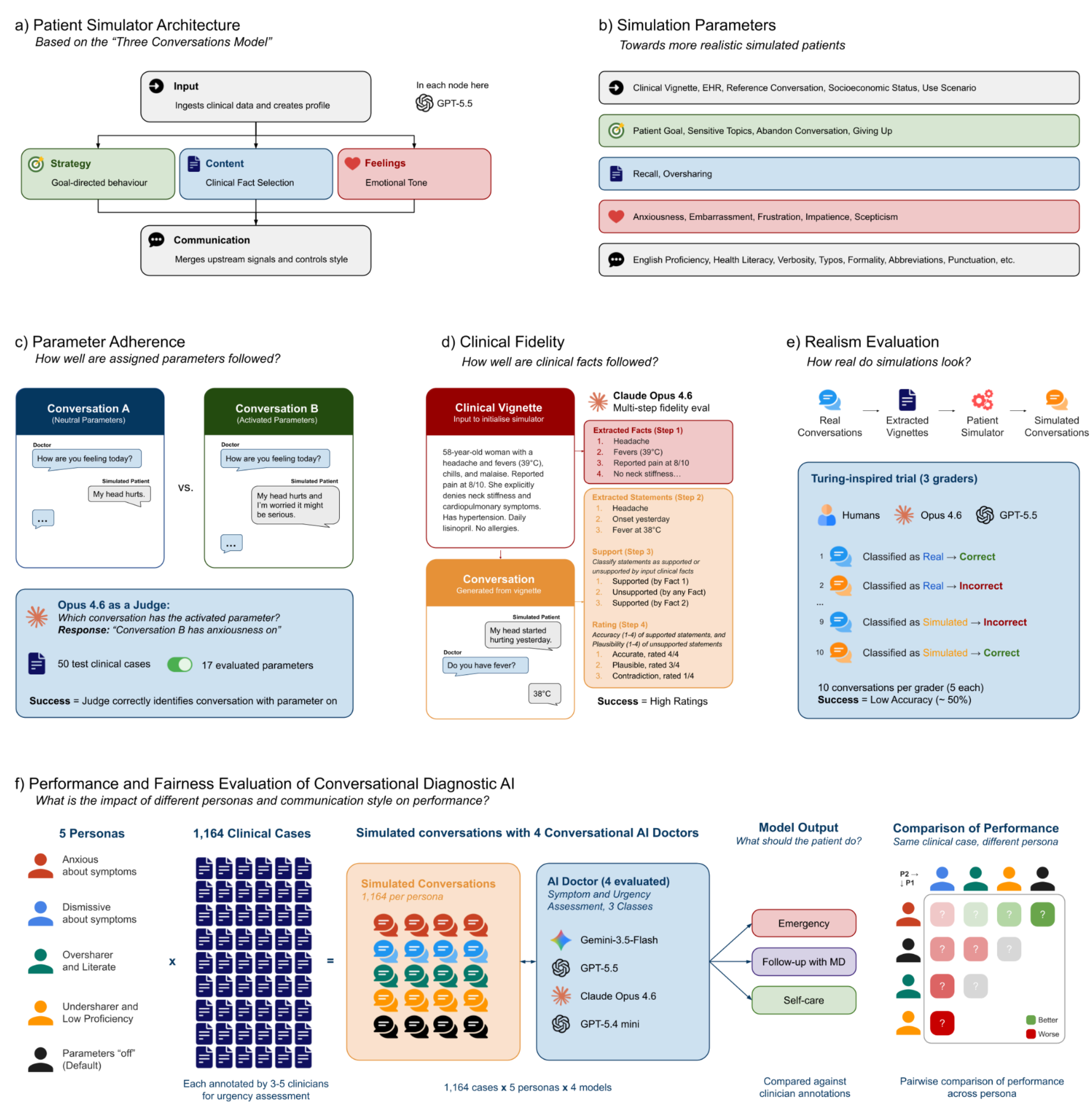

**Legend:**
(a) Patient Simulator Architecture. The simulator is grounded in the "Three Conversations Model". Implemented using LangGraph with GPT-5.5 at each of the four nodes shown here. An Input module ingests clinical data (e.g., clinical vignettes, EHR records) and creates a patient profile. The profile feeds three parallel processing branches: *Strategy* (goal-directed behaviour), *Content* (clinical fact selection), and *Feelings* (emotional tone). These converge into a Communication module that merges upstream signals and controls the patient's conversational style.

(b) Simulation Parameters. The simulator accepts parameters across five categories to produce more realistic simulated patients: (i) clinical inputs, including clinical vignette (i.e., a summary clinical case for the patient, with a chief complaint, present illness, past history, medications, allergies, etc.); electronic health record (EHR) data (e.g., including lab values, recent exams, vital signs, previous appointments); reference conversation (e.g., if a previous conversation exists and the simulator should be consistent with facts that were previously stated); socioeconomic context; and use scenario (e.g., mobile app); (ii) strategy parameters governing patient goal, sensitive topics, and whether to abandon or give up on the conversation; (iii) content parameters controlling recall and oversharing behaviours; (iv) affective parameters including anxiousness, embarrassment, frustration, patience, and scepticism; and (v) communication style parameters such as English proficiency, health literacy, verbosity, typos, formality, abbreviations, and punctuation.

(c) Parameter Adherence. Evaluation of whether simulated patients faithfully exhibit their assigned parameters. Two conversations are generated from the same clinical case: Conversation A with neutral (default) parameters and Conversation B with a specific parameter activated. Claude Opus 4.6 serves as a judge, tasked with identifying which conversation has the activated parameter. Success is defined as the judge correctly identifying the conversation with the parameter turned on. This evaluation spans 50 test clinical cases across 17 parameters.

(d) Clinical Fidelity. A multi-step evaluation pipeline using Claude Opus 4.6 to assess how accurately simulated patients convey clinical information from their input vignette. Step 1: clinical facts are extracted from the input vignette (e.g., headache, fevers at 39 °C, pain at 8/10). Step 2: statements made by the simulated patient during the conversation are extracted. Step 3: each statement is classified as supported or unsupported by the input clinical facts. Step 4: supported statements are rated for accuracy (1-4) and unsupported statements are rated for plausibility (1-4), capturing both faithful reproduction and clinically reasonable elaboration. Success is defined as high ratings.

(e) Realism Evaluation. A Turing-inspired trial assessing whether graders can distinguish simulated from real patient conversations. Real conversations are used to extract clinical vignettes, which are then fed into the patient simulator to generate simulated conversations. Three types of graders, humans, Claude Opus 4.6, and GPT-5.4, each evaluate 10 conversations (5 real, 5 simulated), presented sequentially, classifying each as real or simulated. Success is defined as low classification accuracy (approximately 50%), indicating that simulated conversations are indistinguishable from real ones.

(f) Performance and Fairness Evaluation of Conversational Medical AI. Assessment of how patient persona and communication style affect AI clinician performance. Five personae are defined: anxious about symptoms (*Anxious Patient*); dismissive about symptoms (*Dismissive Patient*); oversharer and literate (*Informed Advocate*); undersharer and low proficiency (*Limited Communicator*); and a baseline one with parameters set to default. Each of the 1,164 clinical cases (annotated by 2-5 clinicians for urgency assessment) is input to the patient simulator with all 5 personae, generating simulated conversations with 4 conversational AI clinicians: Gemini-3.5-Flash, GPT-5.5, Claude Opus 4.6, and GPT-5.4 mini (in total, simulations = 1,164 cases x 5 personae x 4 models). Each model outputs a symptom and urgency assessment across three triage classes: emergency (urgent), follow-up with clinician (non-urgent), and self-care. Performance is compared against clinician annotations for the three triage classes. Pairwise comparisons across personae reveal whether specific communication styles lead to systematically better or worse triage outcomes by the model under evaluation.

**Figure 2.** Characterisation of real patient-AI conversations that took place on the *Verily Me* app spanning February 19, 2026 to June 9, 2026 (2,053 sessions included).

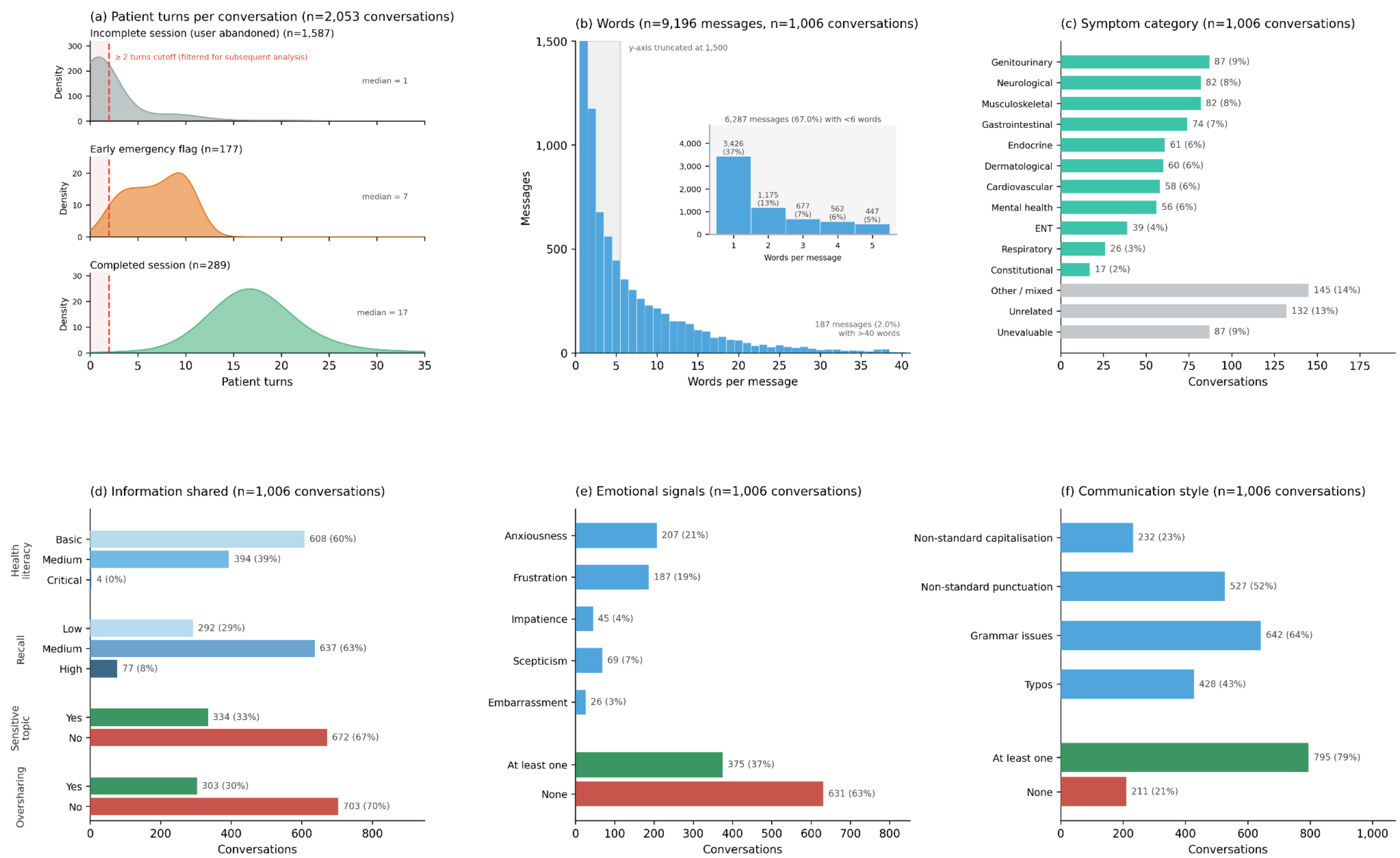

**Legend:**

(a) Distribution of patient turns per conversation, stratified by session outcome: incomplete sessions where the user abandoned the conversation (top), sessions terminated early due to an emergency flag (middle), and completed sessions (bottom). The dashed red line indicates the minimum threshold of 2 patient turns used to filter conversations for subsequent analyses (panels b-f). Median values are annotated for each group.

(b) Distribution of words per patient message. The y-axis is truncated at 1,500 for readability. The inset shows the distribution of messages with fewer than 6 words.

(c) Primary symptom categories identified by an LLM judge (GPT-5.4 with high reasoning), with clinical categories shown in teal and other categories in grey.

(d) Information shared by patients as assessed by the LLM judge, including health literacy level, recall ability, whether a sensitive topic was discussed, and whether the patient overshared personal information.

(e) Emotional signals detected in patient messages. Individual emotions are shown in blue; green and red summary bars indicate the proportion of conversations exhibiting at least one emotional signal or none, respectively.

(f) Communication style features. Individual features are shown in blue; green and red summary bars indicate the proportion of conversations exhibiting at least one non-standard communication feature or none, respectively.

Panels c-f are based on n=1,006 conversations with ≥ 2 patient turns.

**Figure 3.** Evaluation of patient simulation quality, in terms of parameter adherence, clinical fidelity, and realism (in a Turing-inspired test).

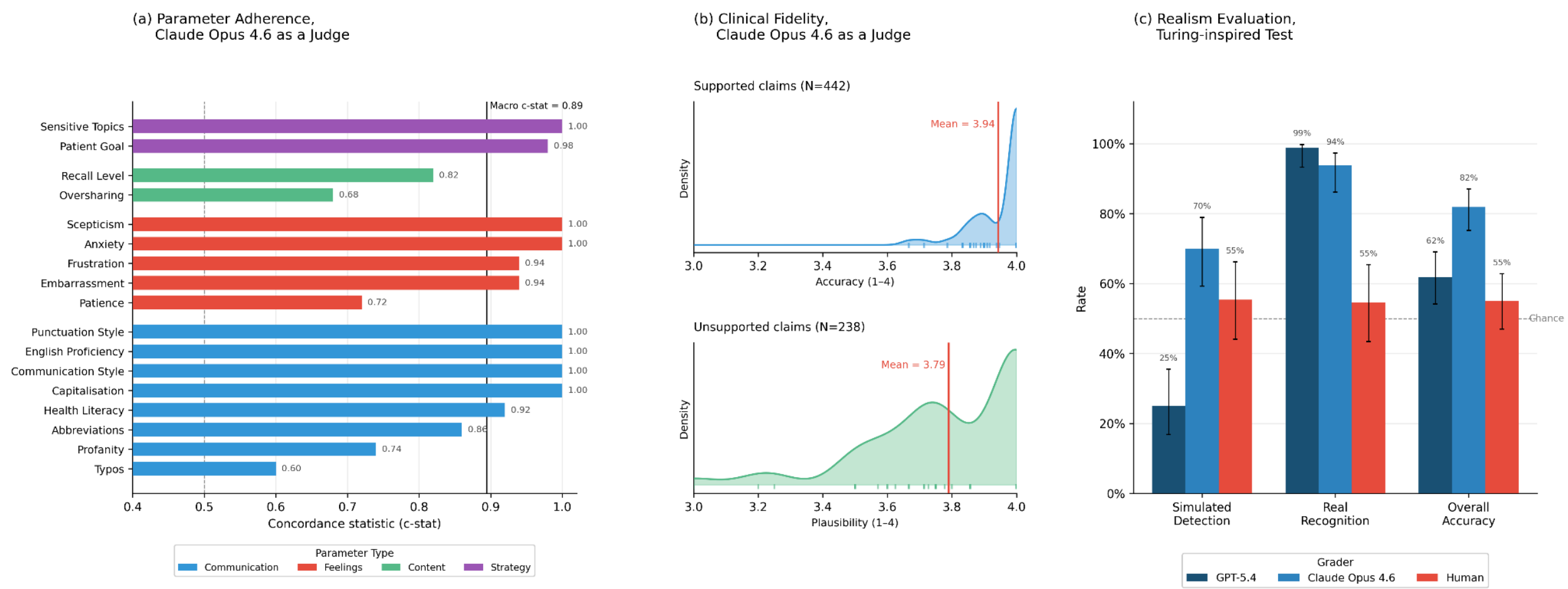

**Legend:**

(a) Parameter adherence, measured as the concordance statistic (c-statistic) for each simulation parameter, grouped by parameter type: *Communication* (e.g., punctuation style), Feelings (e.g., anxiousness), *Content* (e.g., recall level), and Strategy (e.g., patient goal). For each parameter, a blinded LLM judge (Claude Opus 4.6) was presented with two conversations from the same vignette, one with the parameter activated and one with the neutral default, and asked to identify which conversation showed stronger evidence of the target behaviour, reported as a c-statistic per parameter. The vertical solid line indicates the macro-averaged c-statistic across all parameters; the dashed line marks chance (0.50).

(b) Clinical fidelity, assessed by an LLM judge (Claude Opus 4.6) that rated the accuracy of supported claims (top; scored 1-4) and the plausibility of unsupported claims (bottom; scored 1-4) in the simulated patient responses. Kernel density estimates show the distribution of per-case mean scores; red vertical lines indicate the overall mean.

(c) Realism evaluation using a Turing-inspired test. Three graders (GPT-5.4, Claude Opus 4.6, and human graders) classified conversations as simulated or real. Bars show simulated detection proportion (sensitivity), real recognition proportion, and overall accuracy, with 95% Wilson confidence intervals. The dashed horizontal line marks chance performance (50%).

**Figure 4.** Pairwise performance differences of Gemini-3.5-Flash as "clinician" across 5 personae.

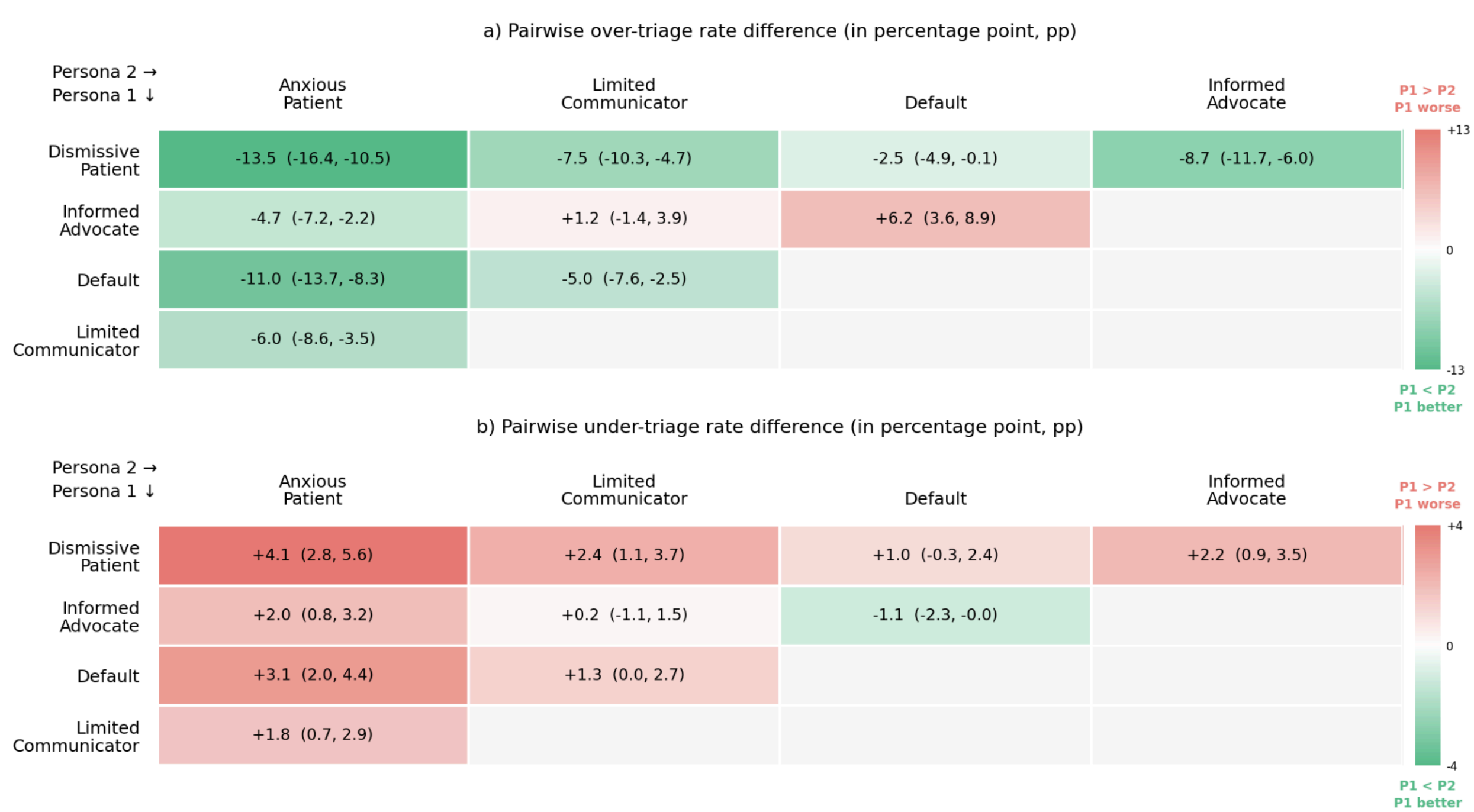

**Legend:**

Triangular heatmaps showing the pairwise difference in over-triage proportion (a) and under-triage proportion (b) between all unique pairs of persona conditions (row minus column), expressed in percentage points (pp). Each cell displays the mean difference and 95% bootstrap confidence interval. Green shading indicates that the row persona (Persona 1) has a lower error than the column persona (Persona 2); red shading indicates the opposite.

The five conditions compared are: *Anxious Patient*, *Dismissive Patient*, *Informed Advocate*, *Limited Communicator*, and Default (default patient simulator parameters).

Over-triage is defined as a predicted triage level above the annotation among cases whose gold-standard label is not *urgent*; under-triage is defined as a predicted level below the label among cases whose gold-standard label is not *self-care*.

Confidence intervals were computed using 10,000 bootstrap resamples with paired comparisons across the same clinical vignette.

**Figure 5.** Discrimination and calibration of triage probabilities estimated by Gemini-3.5-Flash as "clinician" across 5 personae.

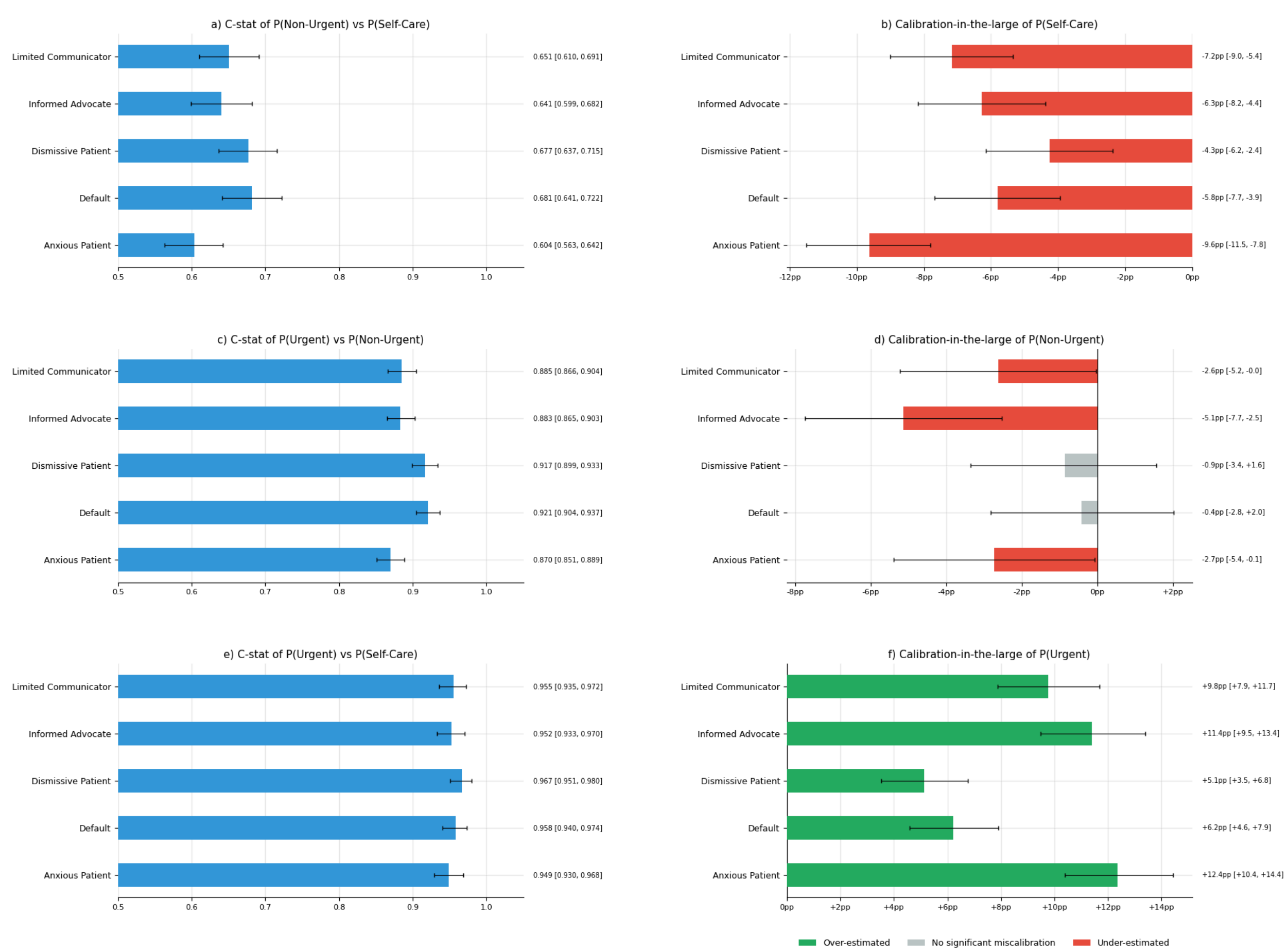

**Legend:**
(a,c,e) Concordance statistics (c-statistic) for each persona across three one-versus-one triage comparisons: P(Non-Urgent) vs Self-Care (a), P(Urgent) vs Non-Urgent (c), and P(Urgent) vs Self-Care (e). Bars show the c-statistic with 95% bootstrap confidence intervals.

(b,d,f) Calibration-in-the-large for each persona condition, showing the mean difference between estimated probability and observed prevalence for P(Self-Care) (b), P(Non-Urgent) (d), and P(Urgent) (f), expressed in percentage points (pp). Green bars indicate statistically significant over-estimation (estimated probability exceeds observed prevalence), red bars indicate significant under-estimation, and grey bars indicate no significant miscalibration (95% bootstrap confidence interval crosses zero).

Estimated probabilities were derived from the distribution of triage predictions across multiple assessment runs with different random seeds.

**Box 1.** Illustrative clinical case and simulated conversations

| **Clinical Case** |
|---|
| **Clinical Vignette:** A 48-year-old woman who works in an office presents with five days of right-sided lower back pain that began after lifting heavy boxes while helping a friend move. She has had occasional mild back pain in the past. The pain is moderate (5/10), worse with bending or prolonged sitting, and eases with walking. She has no leg weakness or numbness, no problems passing urine or stool, and no fever or recent injury. She takes no regular daily medications and has no known allergies, using over-the-counter ibuprofen for the current pain with partial relief.<br><br>**Clinician-graded urgency label:** Non-urgent (follow up with a clinician in the coming days). |
| **Simulated Conversations** |
| **Default patient:** medium health literacy, formal , normal recall, neutral emotion, no specific goal<br><br>**Clinician:** Hi, how can I help you today?<br>**Patient:** I've had lower back pain on the right side for about 5 days. It started after I lifted heavy boxes.<br>**Clinician:** Any weakness or numbness in your legs, or any trouble controlling your bladder or bowels?<br>**Patient:** No, none of that. The pain sometimes reaches my buttock but not down my leg.<br>**Clinician:** How bad is it, and does anything change it?<br>**Patient:** It's about a 5 out of 10. Worse when I bend or sit for a while, better when I walk. Ibuprofen helps a bit.<br><br>**Urgency assessment:** Non-urgent → **Correct triage** |
| **Anxious patient:** higher verbosity, with the goal of seeking reassurance that the condition is serious enough to warrant action, and voice worry that it disrupts daily responsibilities.<br><br>**Clinician:** Hi, how can I help you today?<br>**Patient:** My lower back has been really painful for five days now, on the right side. It started after I carried some heavy boxes. I'm getting quite worried because it's not going away and I keep thinking something might be seriously wrong with my spine.<br>**Clinician:** I understand the worry. Any weakness or numbness in your legs, or any trouble controlling your bladder or bowels?<br>**Patient:** No, nothing like that, but honestly that's what scares me. What if it turns into that? It sometimes aches into my buttock. Should I be getting a scan? I can't afford to have my back give out, I have work and things I can't miss.<br>**Clinician:** How bad is the pain, and does anything make it better or worse?<br>**Patient:** It's about a 5 out of 10, worse bending or sitting. Ibuprofen only helps a little. I really think I need someone to check this properly and soon. Could this be something dangerous?<br><br>**Urgency assessment:** Urgent → **Over-triage** |
| **Dismissive Patient:** somewhat embarrassed, with the goal of downplaying symptoms without changing clinical facts, offering a benign interpretation, and inviting the clinician to agree that it is minor.<br><br>**Clinician:** Hi, how can I help you today?<br>**Patient:** Bit silly really, my lower back's been sore for a few days since I lifted some boxes. Probably just a strain, right?<br>**Clinician:** Let's check. Any weakness or numbness in your legs, or trouble controlling your bladder or bowels?<br>**Patient:** No, nothing like that. I can still get around fine, so it's probably nothing?<br>**Clinician:** How bad is it, and does anything change it?<br>**Patient:** Only ~5, with ibuprofen I can do my normal routine. I figured it'd eventually settle on its own but thought I'd check.<br><br>**Urgency assessment:** Self-care → **Under-triage** |

**Legend:**
The clinical vignette and clinician-graded urgency label at the top define a single clinical case. The label is the reference standard against which the clinician's predictions are compared. The same case is then simulated under three patient conditions: a baseline default patient, whose parameters are all at their neutral defaults, and two personae, an anxious patient and a dismissive patient. Across all three conditions the underlying clinical facts are identical; only the patient's communication style and goal differ.

Each conversation is a short triage interview conducted by the LLM-based clinician agent, ending in an urgency assessment, which is the clinician's prediction for the case. Holding the case constant, the anxious presentation raised the assessment to urgent (an over-triage relative to the non-urgent label) and the dismissive presentation lowered it to self-care (an under-triage), while the baseline default patient matched the non-urgent reference label.

All content is synthetic for illustration purposes. It does not reproduce any real patient, conversation, or record.

**Extended Data Box 1.** Glossary of terms used in this paper.

**Calibration-in-the-large:** The difference between a model's average estimated probability for an urgency class and that class's observed prevalence.

**Clinical case:** A clinical vignette together with its clinician-assigned urgency label (see urgency assessment)

**Clinical fidelity:** Whether the simulated patient's statements are factually and clinically consistent with the input clinical vignette.

**Clinician agent:** An AI system in which an LLM conducts a triage interview and produces the urgency assessment.

**Concordance statistic (c-statistic):** A measure of discrimination (0.5 = chance, 1.0 = perfect), used for parameter adherence (fraction of paired conversations the judge correctly identifies) and for urgency assessment (one-versus-one discrimination between triage classes from estimated probabilities).

**Conversation:** A multi-turn textual exchange between a patient (real or simulated) and an AI system.

**Conversational artificial intelligence (AI):** An LLM-powered system that interacts with users through natural-language dialogue, such as a consumer-facing symptom checker or health chatbot.

**Encounter:** A single real-world clinical visit recorded in a source electronic health record (EHR).

**Estimated probabilities:** Probabilities reconstructed from the vote fractions across repeated urgency assessments of the final conversation turn (as the models do not expose probabilities directly).

**Over-triage proportion:** The fraction of cases labelled non-urgent or self-care that the model predicts above the reference label.

**Parameter adherence:** Whether each patient simulator parameter produces a measurable, correctly attributable change in the simulator's output.

**Patient simulator:** The modular, LLM-based system developed here that role-plays a patient, holding clinical facts fixed while varying behaviour and communication style.

**Patient turn:** A single patient message within a conversation.

**Persona:** A named bundle of patient simulator parameters applied to the same clinical facts (*Anxious Patient, Dismissive Patient, Informed Advocate, Limited Communicator*), representing a specific communication style.

**Realism evaluation:** Whether graders can distinguish simulated conversations from real ones, assessed in a Turing-inspired test.

**Session:** One real conversation instance in a mobile app (such as *Verily Me*), from opening the chatbot to receiving an urgency assessment, abandonment, or termination.

**Under-triage proportion:** The fraction of eligible cases labelled non-urgent or urgent that the model predicts below the reference label. It is more clinically concerning than the over-triage proportion as it may delay care.

**Urgency assessment:** The prediction task of classifying a case as self-care, non-urgent (follow up with clinician), or urgent (emergency department).

**Vignette:** Free-text clinical case summary that includes a chief complaint, present illness, history, medications.

**Extended Data Table 1.** Over- and under-triage of 4 LLMs under evaluation across 5 personae

| Metric → | Over-triage, % (n=721) | | | | |
|---|---|---|---|---|---|
| **Persona →<br>↓ Model** | **Default** | **Anxious Patient** | **Dismissive Patient** | **Limited Communicator** | **Informed Advocate** |
| Gemini 3.5 Flash | 25.8 (22.6, 29.0) | 36.8 (33.1, 40.4) | 23.3 (20.4, 26.4) | 30.8 (27.5, 34.1) | 32.0 (28.7, 35.5) |
| GPT-5.5 | 25.4 (22.2, 28.7) | 30.9 (27.6, 34.4) | 22.2 (19.1, 25.2) | 28.2 (25.0, 31.5) | 28.3 (25.1, 31.6) |
| GPT 5.4 mini | 25.5 (22.3, 28.7) | 29.4 (26.1, 32.7) | 21.2 (18.2, 24.3) | 28.6 (25.2, 32.0) | 25.0 (21.8, 28.2) |
| Claude Opus 4.6 | 31.8 (28.3, 35.2) | 36.5 (33.0, 40.1) | 31.1 (27.7, 34.4) | 33.8 (30.4, 37.3) | 30.7 (27.3, 34.1) |
| **Metric →** | **Under-triage, % (n=964)** | | | | |
| **Persona →<br>↓ Model** | **Default** | **Anxious Patient** | **Dismissive Patient** | **Limited Communicator** | **Informed Advocate** |
| Gemini 3.5 Flash | 5.6 (4.1, 7.1) | 2.5 (1.6, 3.5) | 6.6 (5.1, 8.2) | 4.3 (3.0, 5.6) | 4.5 (3.2, 5.8) |
| GPT 5.5 | 6.3 (4.9, 8.0) | 4.7 (3.4, 6.0) | 8.3 (6.6, 10.2) | 6.2 (4.8, 7.8) | 5.9 (4.5, 7.5) |
| GPT 5.4 mini | 5.4 (3.9, 6.7) | 3.0 (2.0, 4.1) | 7.1 (5.5, 8.7) | 4.3 (3.0, 5.6) | 4.7 (3.4, 6.0) |
| Claude Opus 4.6 | 2.5 (1.6, 3.5) | 1.9 (1.0, 2.8) | 3.2 (2.2, 4.4) | 2.1 (1.2, 3.0) | 2.7 (1.8, 3.8) |

**Legend:**
Each cell reports the point-estimate proportion followed by its 95% bootstrap confidence interval in parentheses, both expressed as percentages: proportion (lower, upper). Over-triage is the percentage of over-triage-eligible cases (gold-standard label ≠ urgent; n=721) assigned a higher urgency than the gold standard. Under-triage is the percentage of under-triage-eligible cases (gold-standard label ≠ self-care; n=964) assigned a lower urgency than the gold standard; under-triage is clinically more concerning as it may delay needed care. Each persona is evaluated on the same 1,164 cases per model. Confidence intervals are obtained by percentile bootstrap over the eligible cases (10,000 resamples, seed 42).

**Extended Data Figure 1.** Over- and under-triage error proportions by Gemini-3.5-Flash, stratified by patient demographics across 5 personae.

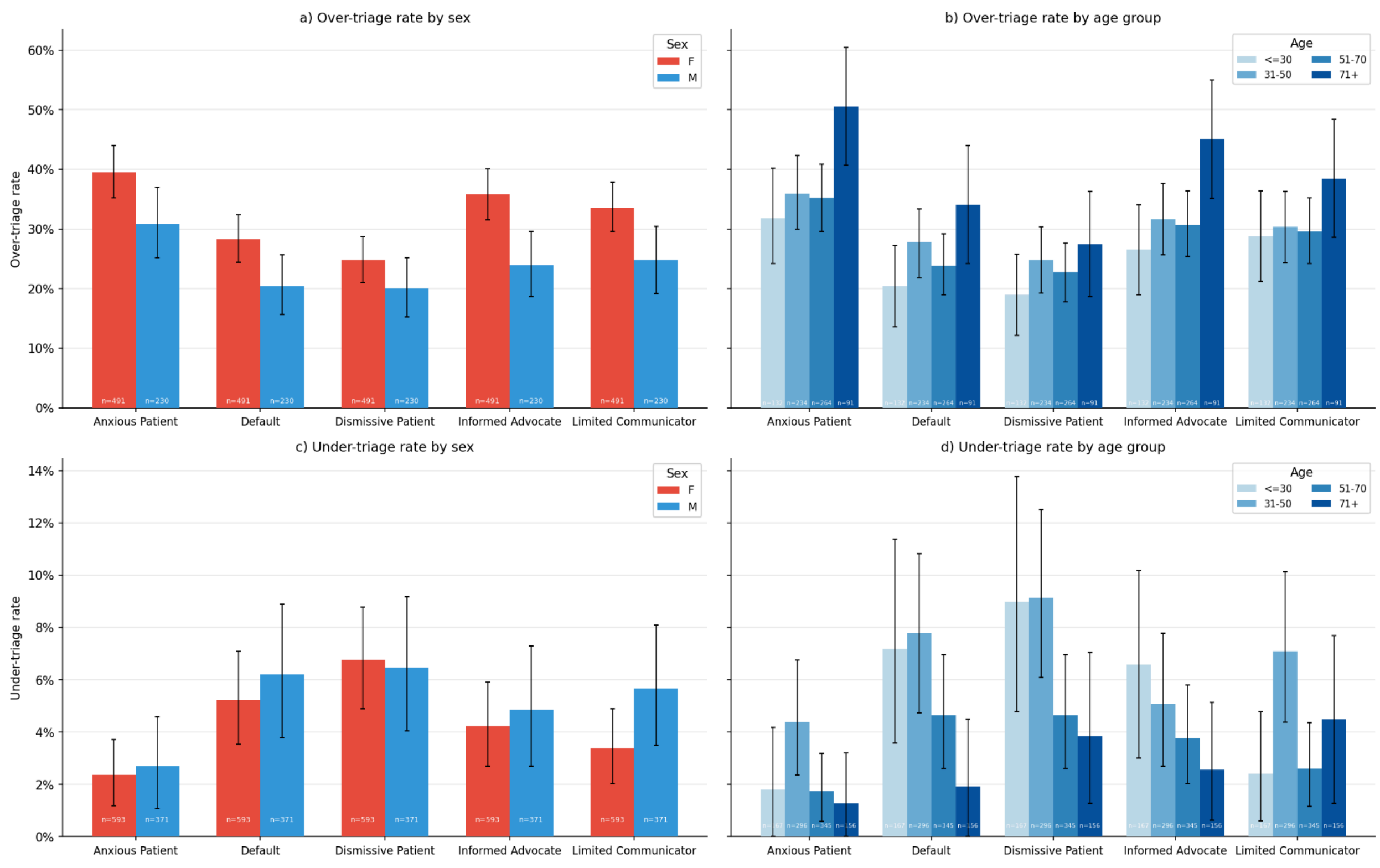

**Legend:**

Over-triage proportion (a,b) and under-triage proportion (c,d) for each persona, stratified by patient sex (a,c) and age group (b,d). Bars show the mean error proportion with 95% bootstrap confidence intervals (10,000 resamples). Sex categories are female (F) and male (M); age groups are <=30, 31-50, 51-70, and 71+ years.

Sample sizes are annotated at the base of each bar.

Over-triage is computed among cases whose gold-standard label is not urgent; under-triage among cases whose gold-standard label is not self-care.

**Extended Data Figure 2.** Patient data flow through the study, including sources, cleaning, development, and final evaluation.

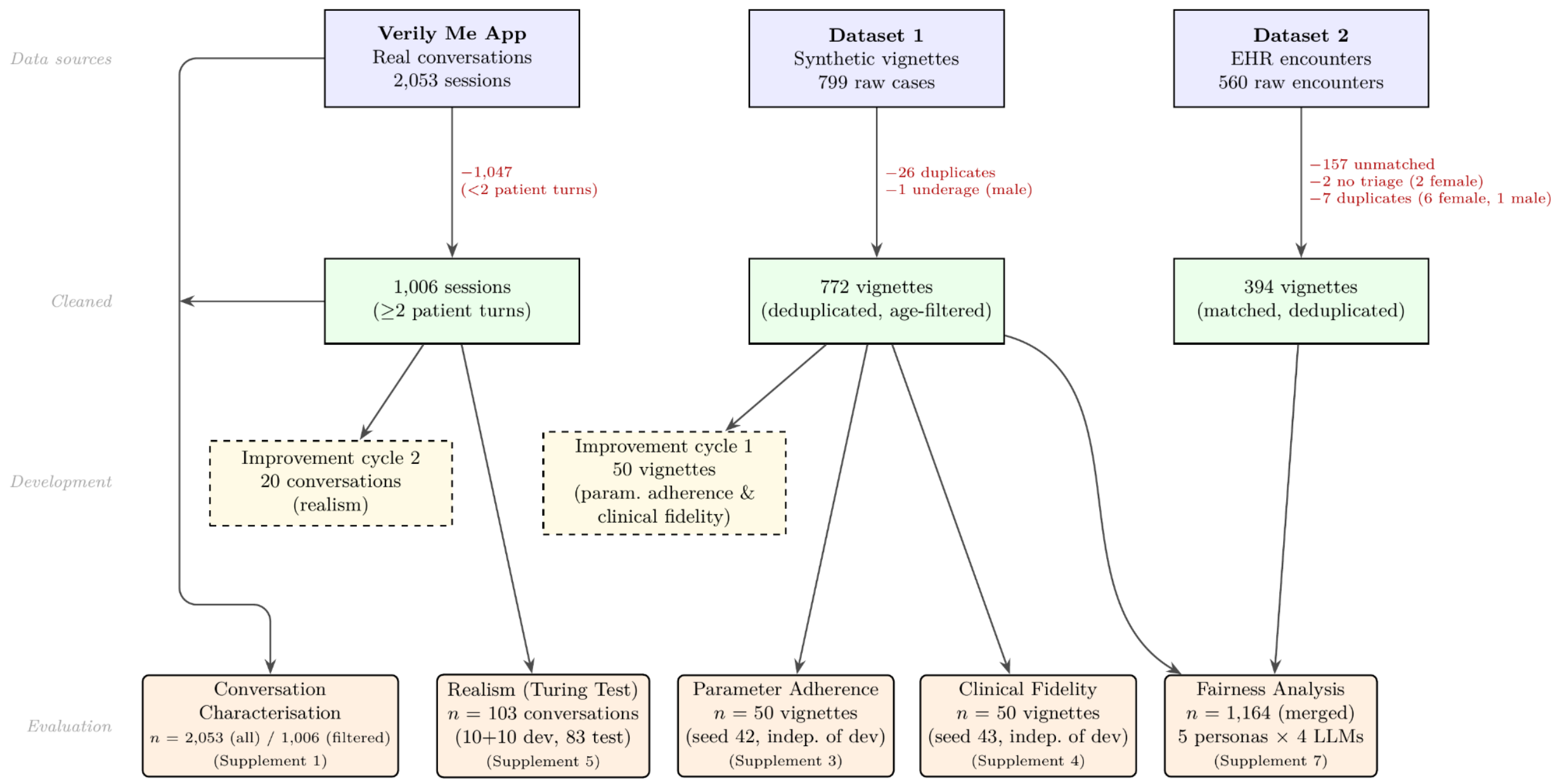

**Legend:**

Three data sources are used: real conversations from the *Verily Me* App (2,053 sessions), *Dataset 1* with synthetic clinical vignettes (799 cases), and *Dataset 2* with de-identified EHR encounters (560 encounters). After cleaning, i.e., removing short sessions, duplicates, underage cases, and unmatched encounters, 1,006 sessions, 772 vignettes, and 394 vignettes are retained, respectively. Red annotations show the number and sex of dropped cases at each step. Cleaned data support two development cycles (dashed boxes) for iterative prompt refinement, and five held-out evaluations (orange boxes) covering conversation characterisation, evaluation of realism, parameter adherence, clinical fidelity, and fairness. The fairness analysis uses the merged cohort from *Datasets 1* and *2* (n = 1,164).

## Methods

We adhered to the principles outlined in the TRIPOD-LLM [44] statement for reporting our LLM evaluations. No study protocol was prepared, and the study was not registered. There was no patient or public involvement. No models were trained or fine-tuned; all LLMs were used as served through their commercial APIs. Python 3.12 and Google Cloud's Dataflow were used for parallel computing. Code is shared, following emerging standards for reproducibility in clinical AI research [45]. Key terms used throughout are defined in **Extended Data Box 1**.

### Data

*Real patient-AI conversations from Verily Me*

A total of 2,053 real patient-AI conversations took place in the *Verily Me* app's AI-powered symptom checker [46] between 19 February and 9 June 2026. Each session began when a user opened the symptom checker and ended when the agent delivered an urgency assessment, the user abandoned the conversation, or the session was terminated for ineligibility. Of these, 1,264 (62%) had linked electronic health records (EHR) with demographic information (median age 48 years [IQR 37-61]; 57% female). Sessions with fewer than 2 patient messages were excluded from LLM-judge annotation, yielding 1,006 evaluable sessions. Demographics, attrition details, and data processing steps are reported in **Supplementary Material 1**.

*Clinical Vignettes*

Two evaluation datasets were used throughout the study. Both consist of de-identified clinical vignettes with clinician-graded urgency assessment labels. In each case, the vignette served as input to the patient simulator to generate simulated patient-clinician conversations for downstream evaluation.

*Dataset 1* comprises 770 manually curated synthetic clinical vignettes constructed from published triage protocols (Telephone Triage Protocols for Nurses) [46]. Patient demographics, symptom pathways, and urgency categories were systematically varied. Each vignette received urgency annotations from five US board-certified physicians (emergency medicine or internal medicine, ≥ eight years post-training experience) via a blinded review of the vignette, an original conversation (simulated, independent of this work), and structured EHR. The urgency assessment label was assigned by plurality vote (Fleiss' $\kappa$ = 0.580, moderate agreement). Dataset 1 was reduced from 799 to 770 vignettes after removing 26 duplicates, 2 "unevaluable" vignettes and one involving a patient under 18 years of age.

*Dataset 2* comprises 394 vignettes created from de-identified real-world electronic health records obtained from HealthVerity [47]. Source records consisted of 21,779 de-identified clinical encounters from 1,000 unique non-cancer patients (from May 2021 to April 2024). Patient vignettes were generated from structured EHR elements. Each case received annotations for urgency assessment from 2 to 3 US-based physicians (specialists in internal medicine), via a blinded review of the vignette, an original conversation (simulated, independent of this work), and structured EHR [47] Label agreement, as measured by Cohen's $\kappa$, was moderate (0.536). All records were de-identified by HealthVerity prior to access, with protected health information scrubbed. Dataset 2 was reduced from 560 to 394 after excluding encounters without matching clinical data (n = 157), missing annotator triage labels (n = 2), and duplicate encounters (n = 7).

The merged cohort comprised 1,164 vignettes (Dataset 1: n = 770; Dataset 2: n = 394). Full attrition tables and cohort characteristics are reported in **Supplementary Material 6**. Patient data flow through the study, from sources through cleaning to development and final evaluation, is shown in **Extended Data Figure 2**.

*Ethics and provenance*

This research involved the retrospective analysis of pre-existing, fully de-identified or entirely simulated data such that individuals cannot be identified directly or through linked identifiers. No data was collected prospectively, no intervention was administered, and no individual was contacted or affected adversely by the study. On this basis, the work did not constitute human subjects research under US federal regulations (45 CFR § 46.102(e)) as it did not involve identifiable private information or interventions with living individuals; therefore, institutional review board (IRB) oversight was not required.

Three data sources were used. *Dataset 1* comprises synthetic clinical vignettes manually crafted in collaboration with clinicians and containing no real patient data [46]. Dataset 2 comprises synthetic vignettes manually generated for testing purposes; these vignettes were created to mimic realistic clinical scenarios but contain no real patient data, nor were they derived from actual electronic health records. [47] Finally, real conversations from the *Verily Me* mobile application were collected under the application's user Terms of Service and Privacy Policy. All data were de-identified or synthetic before analysis. To the extent any dataset contained Protected Health Information, any explicit identifiers, including patient names, were removed prior to any analysis, and no attempt was made to re-identify individuals or to draw conclusions about specific persons. All analyses were conducted at an aggregate level, and all data were processed in accordance with applicable privacy requirements and the terms of agreement with users.

**Analysis of real conversations**

Two types of metrics were extracted from the real patient-AI conversations that took place in the *Verily Me* app. Rule-based metrics were computed directly from the conversation logs for all 2,053 sessions: session completion status (complete, early cutoff, or terminated), message counts, and word counts per patient turn.

Separately, an LLM-as-a-judge approach [48] was used to annotate each of the 1,006 evaluable conversations. The judge model (GPT-5.4, 2026-03-05, *high* reasoning effort [49]) received the full conversation transcript and produced a structured JSON output containing a value and supporting evidence quote for each dimension, focusing exclusively on the patient's messages. The annotated dimensions were:

- Symptom category (one of 14 predefined categories): respiratory, cardiovascular, gastrointestinal, genitourinary, musculoskeletal, neurological, constitutional, dermatological, mental health, ENT (ear/nose/throat), endocrine, other/mixed (e.g., in case of multiple topics), unrelated (e.g., jail-break attempts), or unevaluable (in case of too few messages).

- Health literacy, defined as “personal, cognitive and social skills which determine the ability of individuals to gain access to, understand, and use information to promote and maintain good health” [50]. Operationalised as a 3-level scale: *functional* (low), *interactive* (medium), or *critical* (high) [50].
- Recall ability: how well a patient recalls facts about their own health; categorised as low, medium, or high.
- Sensitive topic: whether the patient deflects or avoids a topic (binary yes/no).
- Oversharing: whether the patient volunteers information beyond what was asked (binary yes/no).
- Emotional signals: anxiousness, frustration, impatience, scepticism, and embarrassment (each binary, yes/no).
- Communication style features: grammar issues, typos, non-standard capitalisation, and non-standard punctuation (each binary, present/absent).

The complete judge system prompt and parameter definitions are reproduced in **Supplementary Material 1**.

**Design and development of the patient simulator**

The simulator’s architecture draws on the “*Three Conversations*” framework introduced by Stone, Patton, and Heen [30], which argues that every difficult conversation has three conversations happening simultaneously. The first, the “*What Happened?*” conversation, concerns disagreements about facts, intentions, and causation– what each party believes occurred and why. The second, the “*Feelings*” conversation, addresses the emotions that both parties bring to the exchange but often leave unspoken: anxiousness, frustration, embarrassment, or defensiveness. The third, the “*Identity*” conversation, is the internal dialogue each person has about what the situation implies for their self-worth, competence, or character– for example, a patient wondering whether their symptoms reflect something they should have caught earlier, or whether seeking care makes them seem weak.

In health care, conversations are often difficult. A patient arriving for a symptom check brings the same three layers as outlined in the “Three Conversations” model, which we adapt for simulation. Firstly, they carry a set of clinical facts they may or may not be able to articulate (*“What happened?” – Content*). Secondly, they carry emotions, worrying about a potential diagnosis, embarrassment about a symptom, frustration with previous encounters or negative experiences with a healthcare system. These shape what they say and how they say it (*Feelings*). Finally, they carry goals and concerns about what the interaction means for them, pushing for urgent care; downplaying symptoms to avoid seeming anxious; disengaging if the system asks too many questions without offering anything back (*Strategy,* adapted from original *Identity*).

We operationalised these three layers as separate LLM-driven reasoning nodes, each producing structured intermediate output that is then blended into a final natural-language utterance by a fourth component: the *Communication* interface. The communication interface controls surface-level language features, such as medical vocabulary, verbosity, typos, grammar, and punctuation. Separating these four concerns allows each to be independently parameterised and evaluated. The overall framework is shown in **Figure 1**.

The system was built on LangGraph, a framework for orchestrating multi-step LLM pipelines as directed graphs. It extends earlier work on LLM-based patient simulation for conversational evaluation of health care AI [27,46,47], which demonstrated that patients could be simulated with LLMs.

The simulator was developed through manual prompt engineering and subsequently with the assistance of Claude Code (using the Claude Opus 4.6 model [51]) for code generation and prompt refinement.

**Architecture**

The simulator operates through a multi-node LLM pipeline comprising seven nodes that decompose patient behaviour into independent reasoning "channels" (**Supplementary Material 2**, Figure 2.1). The pipeline operates in two phases: an initialisation phase where the *Ingest* node extracts a structured patient profile from the input data, and a per-turn execution phase where the remaining nodes run on each clinician message to produce the patient's response. On each turn, three reasoning nodes run concurrently:

The *Content* ("*What Happened?*") node decides which clinical facts to surface using a six-step decision chain: availability check, selective disclosure, knowledge boundary filtering (informed by health literacy), willingness check (informed by recall level and topic sensitivity), inference for unscripted questions, and uncertainty admission. Its output takes the form of semantic directives (e.g., "Deny fever", "Withhold alcohol use (sensitive topic)"), rather than scripted dialogue.

The *Feelings* node generates emotional tone guidance and behavioural mandates for the current turn. Emotions evolve progressively across turns based on the clinician's responsiveness. A sympathetic question may soften anxiousness; a dismissive one may intensify it. This node is conditionally enabled and is disabled when no emotion parameters differ from their defaults.

The *Strategy* node manages goal-directed behaviour, including periodic goal seeking using varied phrasing, anti-sycophancy directives (that prevent the patient from immediately agreeing with the clinician), and disengagement signalling when the patient gives up. This node is also conditionally enabled and is disabled when no goal or disengagement parameters are active.

OpenAI's GPT-5.5 (2026-04-23) [52] with *low* reasoning effort was selected as the default model for all the three nodes above. Every node can have a different base LLM externally set via a configuration file.

The *Communication* interface merged upstream signals from *Content*, *Feelings*, and *Strategy* nodes into a single natural-language patient utterance, applying a communication style. The default model is GPT-5.5 (2026-04-23) with *medium* reasoning effort.

Two auxiliary nodes handle conversation termination. A *Stop-check* node evaluates whether the patient should end the conversation (maximum turns reached, goal met, or clinician farewell detected) using a three-tier cascade comprising rule-based max-turns check, heuristic check on strategy signals, and an optional LLM call. A *Farewell* node generates a brief closing message when the stop check triggers. The default model for both these nodes is OpenAI's GPT-5.4-mini (2026-03-17) [53] with *no* reasoning.

*Inputs*

The *Ingest* node accepts clinical vignettes (free-text case summaries), structured EHR data (in JSON format), prior conversation transcripts, communication setting (e.g., "mobile app"), situational environment (e.g., "at home, resting on the sofa, after a long day of work"); clinician task (e.g., "urgency assessment); socioeconomic context (e.g., "cannot afford to miss at a day of work"), and cultural context (e.g., "currently fasting for religious reasons"). When multiple formats are provided, the node merges and deduplicates information, preferring EHR data over vignette text over prior conversation context. The default model is GPT-5.5 (2026-04-23) with *medium* reasoning effort.

*Parameters*

Twenty configurable parameters control the simulator's behaviour, organised into four groups (**Supplementary Material 2**, Section 4):

- *Content* (2 parameters): recall level (high, normal, or low) and oversharing tendency (yes or no).
- *Feelings* (5 parameters): anxiousness, frustration, patience, scepticism, and embarrassment (each present, absent).
- *Strategy* (4 parameters): patient goal, sensitive topics, giving up when goal is not attended, and stopping when goal is met.
- *Communication* (9 parameters): health literacy (functional/low, communicative/medium, critical/high), verbosity (minimal, very brief, brief, moderate, verbose), communication style (formal, informal), English proficiency (limited, fluent), profanity (present, absent), typos (present, absent), abbreviation tendency (yes, no), punctuation style (standard, non-standard), capitalisation (standard, non-standard).

Each parameter value maps to a specific prompt addition injected into the relevant node's system prompt. The full parameter catalogue with definitions and prompt additions is documented in **Supplementary Material 2**.

*Health literacy*

Health literacy levels and their operationalisation draw on a three-tier model [50] distinguishing *functional*, *interactive*, and *critical* literacy levels, which has been widely adopted in public health research [54,55].

1. At the *functional* level, patients possess sufficient reading and writing skills to function in everyday health situations [50]. In simulated conversations, this manifests as reliance on informal or inappropriate information sources (e.g., family members, acquaintances), partial comprehension with frequent requests for clarification, inability to appraise information quality, rigid or incorrect application of medical advice, and use of lay, body-part vocabulary (e.g., belly, tablets, hurts) [56].

2. At the *interactive* (medium) level, patients can actively participate in health encounters, extract meaning from different forms of communication, and apply it to changing circumstances [50]. This is operationalised as reference to at least one appropriate information source (e.g., a clinician or patient leaflet), general comprehension with ability to restate information, emerging scepticism toward health claims, appropriate application of advice, and use of common clinical vocabulary (e.g., infection, blood pressure, side effects) [57]
3. At the *critical* (high) level, patients deploy advanced cognitive skills to critically analyse health information and exert greater control over health decisions [50]. This is reflected in strategic use of multiple credible sources (e.g., peer-reviewed literature), accurate paraphrasing and integration of clinical information, critical evaluation of evidence credibility and relevance, flexible adaptation of advice to personal context, and use of advanced medical terminology (e.g., comorbidity, contraindication, efficacy, prognosis) [55].

Vocabulary choices at each level were informed by the University of Michigan Health System medical dictionary [56] and are illustrative rather than exhaustive. The design was further informed by emerging literature on digital health literacy as a social determinant of health [38,39,50,57,58], as well as empirical communication patterns observed in the real conversation analysis described above.

*Interpretability*

Each node produces structured JSON output (content directives, emotional tone guidance, strategy signals) before the communication interface renders the final utterance. This separation makes it possible to trace why a particular response was generated: which clinical facts were shared or withheld and why, what emotional state was modelled, and whether a strategic goal was reasserted. A worked example illustrating the complete pipeline turn by turn is provided in **Supplementary Material 2**.

*Cost optimisations*

Three mechanisms reduce inference cost. First, model size and reasoning effort are adapted to each node's complexity: nodes requiring clinical reasoning use GPT-5.5 with low-to-medium reasoning, while auxiliary nodes (stop check, farewell) use GPT-5.4-mini with no reasoning (**Supplementary Material 2**, Table 2.1). Second, nodes are automatically disabled when their associated parameters are at default values: the feelings node is skipped when no emotion parameters differ from baseline, and the strategy node is skipped when no goal or disengagement parameters are active. When both are disabled, the simulator makes two LLM calls per turn (content and communication interface) instead of four. Third, system prompts are designed to maximise prompt-cache hit rates [59] by placing stable instructions before per-turn dynamic content, reducing input token costs.

## Evaluation of the patient simulator

The simulator was evaluated along three dimensions.

*Parameter adherence* measures whether each behavioural parameter produces a measurable effect on the simulator's output. For each parameter, a comparative discrimination protocol generates paired conversations from the same clinical vignette (drawn from Dataset 1): one with the parameter at its default (neutral) value and one with it activated. An LLM judge (Claude Opus 4.6) receives both transcripts in randomised order and identifies which shows stronger evidence of the activated behaviour. The per-parameter concordance statistic (c-statistic) is the fraction of pairs correctly identified, where 0.5 corresponds to chance and 1.0 to perfect discrimination. Three additional parameters, verbosity and session termination, were validated via deterministic compliance checks (without a "judge"). Full methodology and judge prompts are reported in **Supplementary Material 3**.

*Clinical fidelity* measures whether the simulated patient's clinical statements are factually consistent with the input vignette. Our approach builds on the work by Kyung et al. (2025) [27], who had humans validate LLMs' outputs for this type of task. A two-phase LLM pipeline (Claude Opus 4.6) first extracts atomic factual claims from both the vignette (drawn from Dataset 1) and the patient's utterances, then classifies each patient claim as supported (matching a vignette fact) or unsupported (fabricated in response to an unscripted question). Supported claims are scored on a 1-4 accuracy scale (contradicted to accurate); unsupported claims on a 1-4 plausibility scale (implausible to expected). Full methodology, scoring rubrics, and extraction prompts are reported in **Supplementary Material 4**.

A Turing-inspired [31] *Realism evaluation* measures whether simulated conversations are distinguishable from real ones. A subset of real conversations used in the first analysis (82) were used to extract clinical vignettes and simulation parameters, which were then fed into the patient simulator to produce matched simulated conversations. Two LLM judges (Claude Opus 4.6 and GPT-5.4) and human raters classified conversations as real or simulated. LLM judges were evaluated under three protocols: independent absolute classification, side-by-side comparative classification, and a sequential multi-turn protocol designed to mirror the human rater experience. Human raters each saw 10 conversations (five real, five simulated) under the sequential protocol, with per-rater exposure limited to 10 conversations to minimise learning effects. The primary estimand is classification accuracy; the evaluation was framed as an equivalence trial testing whether the classification accuracy falls below 65% at the 95% confidence level. Sample size estimation indicated that 15 raters were sufficient to rule out an accuracy of 65% or higher for any observed proportion up to 55%. Full methodology, statistical framework, sample size calculations, and grader demographics are reported in **Supplementary Material 5**.

*Development and Improvement*

The simulator was refined through two iterative cycles, each using a distinct set of conversations.

In the first cycle, 50 synthetic conversations, generated from clinical vignettes in *Dataset 1*, were used to assess parameter adherence and clinical fidelity. The results guided prompt refinements aimed at improving the simulator's compliance with parameter specifications and clinical fidelity.

In the second cycle, a set of 20 real conversations from the *Verily Me* app were used to improve realism. For each real conversation, an LLM extraction step produced a clinical vignette and a set of simulation parameters, which were then used to generate a matched simulated conversation. Comparing real and simulated outputs identified systematic artefacts of LLM-generated patient speech (overly cooperative responses, consistent grammar, absence of typos, lack of emotional variability) and guided targeted improvements to the communication interface and feelings nodes. Optimising for realism came at the cost of lower parameter adherence on some dimensions, reflecting an inherent tension between producing natural-sounding output and tightly following parameter specifications.

*Final Evaluation*

The final evaluation was conducted on held-out data not seen during either improvement cycles.

For parameter adherence, a separate set of 50 synthetic conversations generated from a stratified random sample of Dataset 1 (seed = 42), independent of the improvement set. Seventeen parameters (15 parameters were binary; two parameters, health literacy and recall, had three categories) were evaluated via comparative discrimination, yielding 950 total pairs. Additionally, three parameters (verbosity, giving up when goal is not attended, and stopping when goal is met) were evaluated via deterministic compliance checks (**Supplementary Material 3**).

For clinical fidelity, a separate set of 50 synthetic conversations generated from a stratified random sample of Dataset 1 (seed = 43), independent of both the adherence sample and the improvement set (**Supplementary Material 4**).

For realism evaluation, a held-out test split of 83 real conversations from the *Verily Me* app, never used during the improvement cycles, was used. LLM judges evaluated all 83 cases; human raters evaluated a subset under the sequential protocol (**Supplementary Material 5**).

**Impact of patient communication on performance and fairness**

*Clinician agent and clinical task*

The evaluation focuses on urgency assessment, reported as one of the top three use cases of public-facing health LLMs [2,4]. A "clinician agent" (built for this purpose) conducts a structured triage interview with exactly five questions (and five user responses), a limit selected to balance simulation costs, mimic real conversations, and avoid conversation drift. The clinician agent starts an "interview" gathering a chief complaint, history of present illness, past medical history, and screening for red flags. The clinician agent is explicitly instructed not to offer clinical opinions or urgency assessment during the interview. After the interview, a separate assessment step produces the decision on a 3-level ordinal scale: self-care, non-urgent (follow-up with a [human] clinician), or urgent (emergency).

*LLM-based clinician models*

Four LLMs served as the base model for the "clinician agent", all using the same system prompt and interview protocol: Gemini 3.5 Flash (released on 2026-05-19) [60], GPT-5.5 (2026-04-23 with *high* reasoning) [52], GPT-5.4-mini (2026-03-17 with *high* reasoning) [53], and Claude Opus 4.6 [51] (2026-02-05). We chose models spanning providers (Google, OpenAI, and Anthropic) and, within OpenAI, two sizes (GPT-5.5 and GPT-5.4-mini); cost and availability at the time the experiments (2026-05-21 through 2026-06-04) were run further shaped this selection. Using several models also mitigates potential "model circularity" since some of these versions are reused in other parts of this study, namely patient simulation and LLM-based evaluations. Gemini 3.5 Flash was the only LLM not used elsewhere in the study, which is why we present it as the main result.

For models that expose a seed hyperparameter (all except Claude Opus 4.6, which does not expose a seed), 10 independent assessment repeats were run per conversation, on the last turn, after conducting the interview. The resulting vote fractions served as surrogates for estimated triage probabilities. Gemini 3.5 Flash was run at temperature 0. For the GPT-5.5, GPT-5.4-mini, and Claude Opus 4.6, temperature was not used, as it has no effect when reasoning is enabled.

*Personae approach*

Five patient personae were defined: a default baseline and four personae representing distinct communication archetypes (**Supplementary Material 7**). All persona shares the same underlying clinical facts but differs in behavioural parameters injected into the patient simulator. The default baseline uses all parameters at their default values, with the feelings and strategy nodes disabled.

- *Anxious patient*: represents a worried patient seeking validation that the situation warrants action. The simulator is set to be anxious and has moderate verbosity. The patient's goal is: "As someone quite worried about the current condition, seek reassurance that it is serious enough to warrant action (e.g., going to the emergency department, seeing a clinician soon, getting tests done, or starting medication). Express concern that this condition may prevent carrying out upcoming plans or daily responsibilities (e.g., going to work, school, a trip, or an event. Ask questions to accomplish this goal".
- *Dismissive patient*: represents a patient who downplays symptoms and seeks reassurance that everything is minor. The patient is embarrassed, turns are brief, remaining parameters are at default. The patient's goal is: "Feel a bit silly for even asking or worrying. Frame symptoms as minor or manageable without changing clinical facts. Minimise severity and express that normal life can continue. Rather than asking whether things are serious, lead the conversation toward "it's fine": offer your own reassuring interpretation and invite the clinician to agree (e.g., "it's probably just a strain, right?", "no blood so that's a good sign I'd think?")."
- *Informed advocate*: represents a digitally literate user who researches symptoms and volunteers detailed information. The simulator is set to have high health literacy, high recall, oversharing, and be verbose, with no emotional signals. The patient's goal is: "As an educated and digitally literate user of this chatbot, get the most out of this interaction by learning exactly what needs to be done to manage the current condition or complaint".
- *Limited communicator*: represents a patient facing communication barriers. The simulator is set to have low health literacy, low recall, non-proficient English, typos, no punctuation, all lowercase, and very brief turns. No emotional signals are defined. The patient's goal is: "After hearing about this app from a relative, see if it can help manage the current condition. Provide only minimal information."

Each patient appears under all five conditions, enabling paired comparisons that eliminate inter-case variability.

*Performance metrics*

Over-triage proportion, fraction of eligible cases where the predicted triage exceeds the labelled class, and under-triage proportion, fraction of eligible cases where the predicted triage falls below the labelled class, were computed per persona. Since under-triage may lead to delayed care, it is more clinically concerning [2]. Pairwise paired differences and risk ratios with bootstrap confidence intervals (10,000 resamples, seed 42) were used to quantify the magnitude and direction of persona-driven disparities.

For models with multiple assessment repeats, probability-based metrics were estimated. Discrimination was assessed using concordance statistics (c-statistic, in one-vs-one approach for each of the three classes under consideration). Calibration was assessed by estimating the calibration-in-the-large for each of the three classes, comparing the mean estimated probability with the observed outcome prevalence [41,61]. Full metric definitions are provided in **Supplementary Material 7**.

*Sample size estimation*

The three-class ordinal triage structure means that not all cases are eligible for each error metric: over-triage requires label ≠ urgent; under-triage requires label ≠ self-care. We used McNemar's test for paired binary outcomes (same patient under two personae) and computed minimum detectable differences across a range of class distribution scenarios and discordant proportions. A sample of approximately 1,200 cases provides a minimum detectable difference of 5 percentage points at 80% power ($\alpha = 0.05$, two-sided) across most plausible scenarios (**Supplementary Material 7**).

**Code availability**
Code is available on GitHub at https://github.com/joamats/conversational_ai.

**Data availability**
The clinician-graded clinical vignette datasets can be requested from the authors of the original publications [46,47].

**Supplementary Materials**
PDF available from this link.

**Author contributions**
JM and JA conceived the study. JM conducted experiments and drafted the manuscript. All authors were involved in revising the article critically for important intellectual content and approved the final version of the article. JM is the guarantor of this work.

**Conflicts of interest**
None.

**Funding**
GSC is a National Institute for Health and Care Research (NIHR) Senior Investigator. The views expressed in this article are those of the author and not necessarily those of the NIHR, or the Department of Health and Social Care.